\documentclass[10pt,twocolumn,letterpaper]{article}

\usepackage{cvpr}              

\usepackage{graphicx}
\usepackage{amsmath}
\usepackage{amssymb}
\usepackage{booktabs}
\usepackage{epsfig}
\usepackage{url}

\usepackage[pagebackref,breaklinks,colorlinks]{hyperref}

\def\eg{{\em e.g.}}
\def\ie{{\em i.e.}}
\def\etal{{\em et al.}}

\usepackage[capitalize]{cleveref}
\crefname{section}{Sec.}{Secs.}
\Crefname{section}{Section}{Sections}
\Crefname{table}{Table}{Tables}
\crefname{table}{Tab.}{Tabs.}

\usepackage[accsupp]{axessibility} 

\def\cvprPaperID{1878} 
\def\confName{CVPR}
\def\confYear{2022}

\begin{document}

\title{Multi-Granularity Alignment Domain Adaptation for Object Detection}

\author{Wenzhang Zhou$^{1}$, Dawei Du$^{2}$, Libo Zhang$^{1,3}$\thanks{Corresponding author (libo@iscas.ac.cn)}, Tiejian Luo$^{1}$, Yanjun Wu$^{3}$\\
$^1$University of Chinese Academy of Sciences, Beijing, China\\ 
$^2$Kitware, Inc., NY, USA\\
$^3$Institute of Software, Chinese Academy of Sciences, Beijing, China\\
{\tt\small \url{https://github.com/tiankongzhang/MGADA}}}
\maketitle

\begin{abstract}
Domain adaptive object detection is challenging due to distinctive data distribution between source domain and target domain. In this paper, we propose a unified multi-granularity alignment based object detection framework towards domain-invariant feature learning. To this end, we encode the dependencies across different granularity perspectives including pixel-, instance-, and category-levels simultaneously to align two domains. Based on pixel-level feature maps from the backbone network, we first develop the omni-scale gated fusion module to aggregate discriminative representations of instances by scale-aware convolutions, leading to robust multi-scale object detection. Meanwhile, the multi-granularity discriminators are proposed to identify which domain different granularities of samples (\ie, pixels, instances, and categories) come from. Notably, we leverage not only the instance discriminability in different categories but also the category consistency between two domains. Extensive experiments are carried out on multiple domain adaptation scenarios, demonstrating the effectiveness of our framework over state-of-the-art algorithms on top of anchor-free FCOS and anchor-based Faster R-CNN detectors with different backbones.
\end{abstract}

\section{Introduction}
Owing to the emergence of deep learning \cite{xu2021artificial}, modern object detection methods \cite{DBLP:journals/pami/RenHG017,DBLP:conf/cvpr/LinDGHHB17,DBLP:conf/iccv/LinGGHD17,DBLP:conf/eccv/LawD18,DBLP:conf/iccv/TianSCH19} have achieved remarkable progress based on large-scale annotated datasets. However, such domain constrained models often fail in new environments without labeled training data. 

To tackle this problem, a feasible solution is to reduce the disparity between label-rich source domain and label-agnostic target domain by \textit{unsupervised domain adaptation} in an adversarial manner \cite{DBLP:conf/icml/GaninL15}. Specifically, the domain discriminator is introduced to identify whether the image is from source domain or target domain; while the object detector learns domain-invariant features to confuse the discriminator \cite{DBLP:conf/cvpr/SaitoUHS19}. However, classic domain adaptation frameworks suffer from scale variations in cluttered background, resulting in limited performance. 
Due to convolution layers with fixed kernels in the network, it is difficult to capture accurate features of objects with various scales and aspect ratios. For small objects, the features are convolved from a large region with too much background; for large objects, convolutions cover a small part and lack global structural information.
\begin{figure}[t]
\centering
\includegraphics[width=0.75\linewidth]{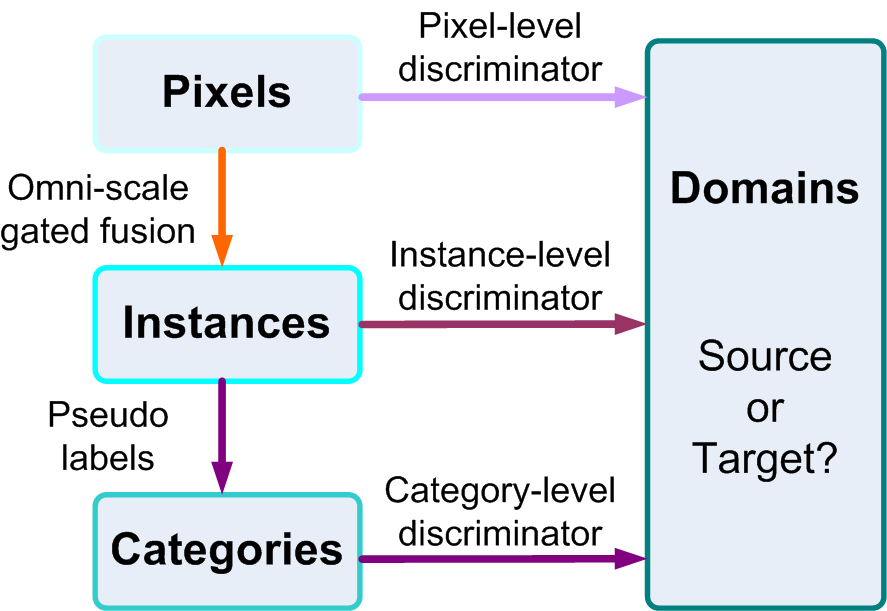}
\caption{Our framework to encode dependencies across multiple granularities including pixel-, instance-, category-level.}
\label{fig:dependency}
\vspace{-6mm}
\end{figure}

On the other hand, for better adaptation in target domain, some researchers employ various feature alignment strategies from different granularity perspectives, \ie, instance-, pixel- and category-level. \textit{Instance-level alignment}~\cite{DBLP:conf/cvpr/Chen0SDG18,DBLP:conf/eccv/LiDZWLWZ20} relies on pooled features of detection proposals to help train the domain discriminator. However, instance-level pooling operation may distort the features of objects with various scales and aspect ratios. In contrast, \textit{pixel-level alignment}~\cite{DBLP:conf/cvpr/KimJKCK19,DBLP:conf/eccv/HsuTLY20} focuses on aligning lower-level features that account for each pixel to handle cross-domain variations of objects and the background. However, there exists a large gap between pixel-level features for different scales of objects with the same category. Recently, \textit{category-level alignment}~\cite{DBLP:conf/cvpr/XuZJW20,DBLP:conf/cvpr/XuWNTZ20} leverages the categorical discriminability in two domains to deal with the hard aligned instances. However, these works pay more attention on the consistency between the image-level and instance-level predictions.

To solve the above issues, we propose a unified multi-granularity alignment based object detection framework by using unsupervised domain adaptation. As shown in Figure \ref{fig:dependency}, we encode the dependencies in different granularity perspectives including pixel-, instance-, and category-levels to align source domain and target domain, which is not a rough combination of previous single-granularity alignment techniques.
To adapt to various instances, our omni-scale gated fusion selects the most plausible convolutions from low-resolution and high-resolution streams to extract the features. Concretely, we first estimate coarse detections as the guidance based on pixel-level backbone feature maps. Then, parallel convolutions are activated to aggregate discriminative representations of instances with similar scales and aspect ratios. In this way, the following object detection head can predict multi-scale objects more accurately.
Meanwhile, we introduce a new category-level discriminator to consider not only the instance discriminability in different categories but also the category consistency between source and target domains. To supervise the category-level discriminator, we assign pseudo labels to important instances with high confidence from object detection.
In summary, we construct the multi-granularity discriminators in three granularities of samples (\ie, pixels, instances, and categories). Thus complementary information in different granularity can support each other and achieve better domain adaptation performance. 

To verify the effectiveness of our method, we conduct comprehensive experiments on different domain adaptation scenarios (\ie, Cityscapes \cite{DBLP:conf/cvpr/CordtsORREBFRS16}, FoggyCityscapes \cite{DBLP:journals/ijcv/SakaridisDG18}, Sim10k \cite{DBLP:/conf/icra/driving17}, KITTI \cite{DBLP:/conf/cvpr/are12}, PASCAL VOC \cite{DBLP:journals/ijcv/EveringhamGWWZ10}, Clipart \cite{DBLP:conf/cvpr/InoueFYA18} and Watercolor \cite{DBLP:conf/cvpr/InoueFYA18}). 
The proposed framework is evaluated on top of anchor-free FCOS \cite{DBLP:conf/iccv/TianSCH19} and anchor-based Faster R-CNN \cite{DBLP:journals/pami/RenHG017} with VGG-16 \cite{DBLP:journals/corr/SimonyanZ14a} and ResNet-101 \cite{DBLP:conf/cvpr/HeZRS16} backbones, achieving state-of-the-art performance on different datasets. For example, our method achieves $43.8\%$ mAP score adapting from the source domain Cityscapes \cite{DBLP:conf/cvpr/CordtsORREBFRS16} to the target domain FoggyCityscapes \cite{DBLP:journals/ijcv/SakaridisDG18} using FCOS \cite{DBLP:conf/iccv/TianSCH19}, which is $3.6\%$ better than the second best method CFA \cite{DBLP:conf/eccv/HsuTLY20}.

\textbf{Contributions}. 1) We propose the multi-granularity alignment framework to encode dependencies across pixel-, instance- and category-level granularities for adaptive object detection, which can be applied in different object detectors. 2) The omni-scale gated fusion module is designed to extract a discriminative representation in terms of objects with different scales and aspect ratios. 3) The category-level discriminator models both instance discriminability in different categories and category consistency between source domain and target domain. 4) Our method achieves the state-of-the-art performance on five domain adaptation applications.

\section{Related Work}
\subsection{Object Detection}
CNNs based object detection methods can be generally grouped into anchor-based and anchor-free frameworks. Anchor-based detectors use a series of anchor boxes with different scales and aspect ratios to generate detection proposals, and then apply a network to classify and regress each candidate object. Faster-RCNN \cite{DBLP:journals/pami/RenHG017} develops the region proposal network (RPN) to generate proposals efficiently. FPN \cite{DBLP:conf/cvpr/LinDGHHB17} introduces a new top-down architecture with lateral connections to capture multi-scale feature maps.
In contrast, anchor-free methods rely on keypoints to represent objects. CornerNet \cite{DBLP:conf/eccv/LawD18} is the pioneering work to detect an object bounding box as a pair of the top-left and bottom-right corners. Recently, FCOS \cite{DBLP:conf/iccv/TianSCH19} leverages the fully convolutional networks to predict labels and bounding box coordinates of each pixel in feature maps. 
In this work, we build our domain adaptation framework on two representative detectors that are widely used in previous domain adaptation methods, \ie, anchor-based Faster-RCNN \cite{DBLP:journals/pami/RenHG017} and anchor-free FCOS \cite{DBLP:conf/iccv/TianSCH19}.

\subsection{Unsupervised Domain Adaptation}
Given the labeled source data and unlabeled target data, unsupervised domain adaptation in object detection attracts the interest of researchers. Ganin and Lempitsky \cite{DBLP:conf/icml/GaninL15} perform domain adaptation for classification networks through standard backpropagation training. Inspired by \cite{DBLP:journals/jmlr/GaninUAGLLML16}, detection networks are optimized by adversarial learning \cite{DBLP:conf/iccv/TsaiSSC19,DBLP:conf/cvpr/ZhangZCYW20}. They apply a domain discriminator to distinguish the feature differences between source and target domains, and a gradient reversal layer to reduce the feature distribution differences between domains. Inoue~\etal~\cite{DBLP:conf/cvpr/InoueFYA18} propose the cross-domain weakly supervised object detection method by fine-tuning the detector on two types of artificially and automatically generated samples. Saito~\etal~\cite{DBLP:conf/cvpr/SaitoUHS19} develop a strong-weak distribution alignment method that adjusts the ability of distribution alignment on local and global levels. Recently, Zheng~\etal~\cite{DBLP:conf/cvpr/Zheng0LW20} generate the attention map with the predicted object categories and sizes to choose regions of objects. In contrast, with the aid of coarse detection, our omni-scale gated fusion module aggregates instance features by scale-aware convolutions to adapt to multi-scale objects in a soft-decision way.

\subsection{Alignment Strategies in Domain Adaptation}
As discussed in the introduction section, to improve upon domain-level alignment, various feature alignment schemes are applied in other finer levels, \ie, instance-level~\cite{DBLP:conf/cvpr/Chen0SDG18,DBLP:conf/eccv/LiDZWLWZ20}, pixel-level~\cite{DBLP:conf/cvpr/KimJKCK19,DBLP:conf/eccv/HsuTLY20} and category-level~\cite{DBLP:conf/iccv/DuTYFXZYZ19, DBLP:conf/cvpr/HuKSC20, DBLP:conf/eccv/PaulTSRC20, DBLP:conf/eccv/WangSZD020, DBLP:conf/cvpr/XuZJW20, DBLP:conf/cvpr/XuWNTZ20}. 

Chen~\etal~\cite{DBLP:conf/cvpr/Chen0SDG18} deal with the domain shift on two levels including image-level (\eg, image style and illumination) and instance-level (\eg, object appearance and size). Li~\etal~\cite{DBLP:conf/eccv/LiDZWLWZ20} propose the spatial attention pyramid network to capture context information of objects at different scales. Kim~\etal~\cite{DBLP:conf/cvpr/KimJKCK19} design the multi-domain-invariant representation learning to encourage the unbiased semantic representation through adversarial learning. Hsu~\etal~\cite{DBLP:conf/eccv/HsuTLY20} propose a center-aware alignment based domain adaptation method to focus on pixel-wise objectness. 

\begin{figure}[t]
\centering
\includegraphics[width=\linewidth]{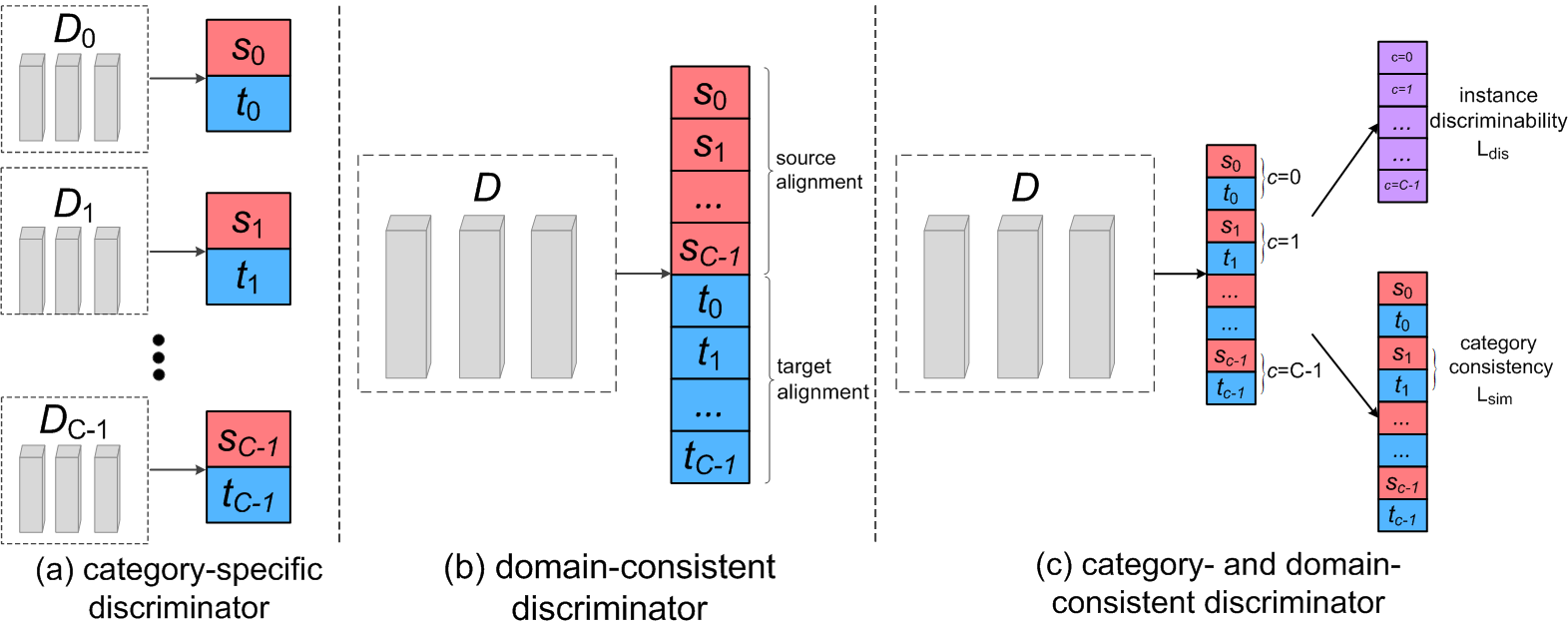}
\caption{Illustration of different category-level discriminators $D$, where $s_c$ and $t_c$ denote the $c$-th category ($c=0,1,\cdots,C-1$) in source domain and target domain respectively. (a) Category-specific discriminators for each category \cite{DBLP:conf/iccv/DuTYFXZYZ19, DBLP:conf/cvpr/HuKSC20, DBLP:conf/eccv/PaulTSRC20}. (b) Domain-consistent discriminator to distinguish different categories within one domain~\cite{DBLP:conf/eccv/WangSZD020}. (c) Our category- and domain-consistent discriminator to consider both instance discriminability in different categories and category consistency between two domains.}
\label{fig:discriminator}
\vspace{-4mm}
\end{figure}

{\noindent \textbf{Category-level alignment.}} 
In terms of category-level alignment, some works \cite{DBLP:conf/iccv/DuTYFXZYZ19, DBLP:conf/cvpr/HuKSC20, DBLP:conf/eccv/PaulTSRC20} design a category-specific discriminator for each category and focus on classification between source and target domains based on pseudo labels (see Figure \ref{fig:discriminator}(a)). It is difficult to learn discriminative category-wise representation among multiple discriminators. Wang~\etal~\cite{DBLP:conf/eccv/WangSZD020} retain one discriminator to distinguish different categories within one domain (see Figure \ref{fig:discriminator}(b)). However, it consider little about the consistency of feature subspaces in the same category across two domains. Besides, a categorical regularization method is developed in ~\cite{DBLP:conf/cvpr/XuZJW20} to locate crucial image regions and important instances to reduce the domain discrepancy. Similarly, Xu~\etal~\cite{DBLP:conf/cvpr/XuWNTZ20} seek for category-level domain alignment by enhancing intra-class compactness and inter-class separability. It builds the graph based on the Euclidean distances between different category prototypes, where the feature subspaces follow the Gaussian distribution. 

In contrast, our category-level discriminator does not rely on the Gaussian distribution assumption but selects important instances to model the subspaces based on an adaptive threshold. Then we model both instance discriminability in different categories and category consistency between two domains (see Figure \ref{fig:discriminator}(c)). Moreover, based on the merits of feature alignment at different levels, our method is a unified domain adaptation framework by taking all the granularities into consideration. 
\begin{figure}[t]
\centering
\includegraphics[width=\linewidth]{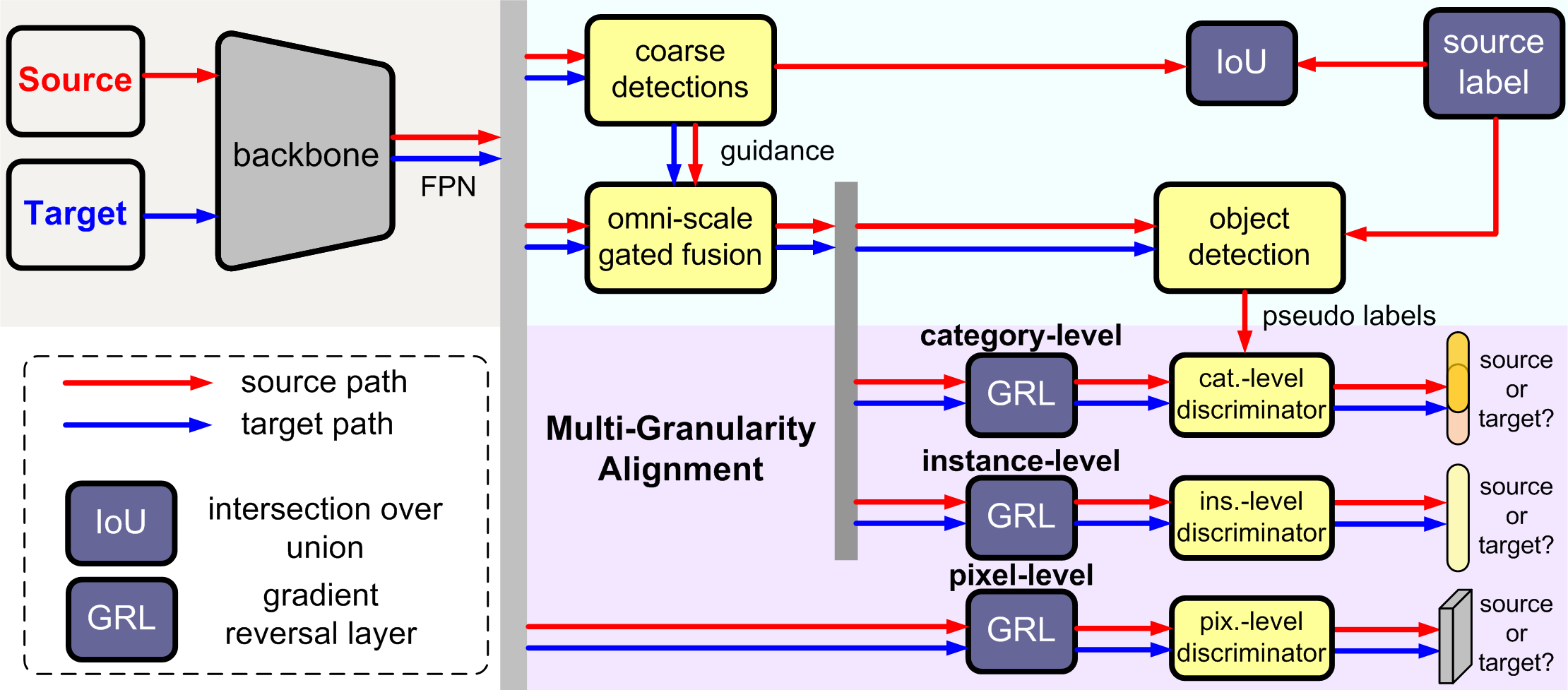}
\caption{Architecture of our domain adaptive object detection network. Note that object detection heads and discriminators have different sizes of outputs with regard to different detectors.}
\label{fig:architecture}
\vspace{-4mm}
\end{figure}
\section{Multi-Granularity Alignment}
As shown in Figure \ref{fig:architecture}, given images from the source domain $s$ and target domain $t$, we first compute the base feature maps using the backbone. Then, the pixel-level features are merged to generate discriminative representations of multi-scale instances by the omni-scale gated fusion module. Based on merged features, the objects can be estimated by the object detection head more accurately. Meanwhile, we introduce the multi-granularity discriminators to distinguish the feature distribution between source and target domains in different perspectives, including pixel-level, instance-level and category-level. 

Notably, our method can be applied in different detectors (\eg, anchor-based Faster-RCNN \cite{DBLP:journals/pami/RenHG017} and anchor-free FCOS \cite{DBLP:conf/iccv/TianSCH19}) and backbones (\eg, VGG-16 \cite{DBLP:journals/corr/SimonyanZ14a} and ResNet-101 \cite{DBLP:conf/cvpr/HeZRS16}). Without loss of generality, we first take FCOS \cite{DBLP:conf/iccv/TianSCH19} as an example, and then explain how our method is applied in Faster-RCNN \cite{DBLP:journals/pami/RenHG017}. For the FCOS detector \cite{DBLP:conf/iccv/TianSCH19}, we extract the last three stages of backbone feature maps and combine them into multi-level feature maps $F^k, k\in\{3,4,5,6,7\}$ using the FPN representation \cite{DBLP:conf/cvpr/LinDGHHB17}. 

\subsection{Omni-Scale Gated Object Detection}\label{sec_detection}
Most previous domain adaptation methods focus on designing discriminators at specific level and attentive regions. However, the point representation in anchor-free models \cite{DBLP:conf/eccv/HsuTLY20,DBLP:journals/corr/abs-2110-00249} is difficult to extract robust and discriminative feature in cluttered background, while the AlignROI operation in anchor based models \cite{DBLP:conf/cvpr/SaitoUHS19,DBLP:conf/eccv/HeZ20} may distort the features of objects with various scales and aspect ratios. 

To solve this issue, we employ the omni-scale gated fusion to adapt to various instances with different scales and aspect ratios. Concretely, with the scale guidance from coarse detections, the most plausible convolutions with different kernels are selected to extract the compact features of instances in terms of object scale. Thus it can be applied in different detectors.

\begin{figure}[t]
\centering
\includegraphics[width=\linewidth]{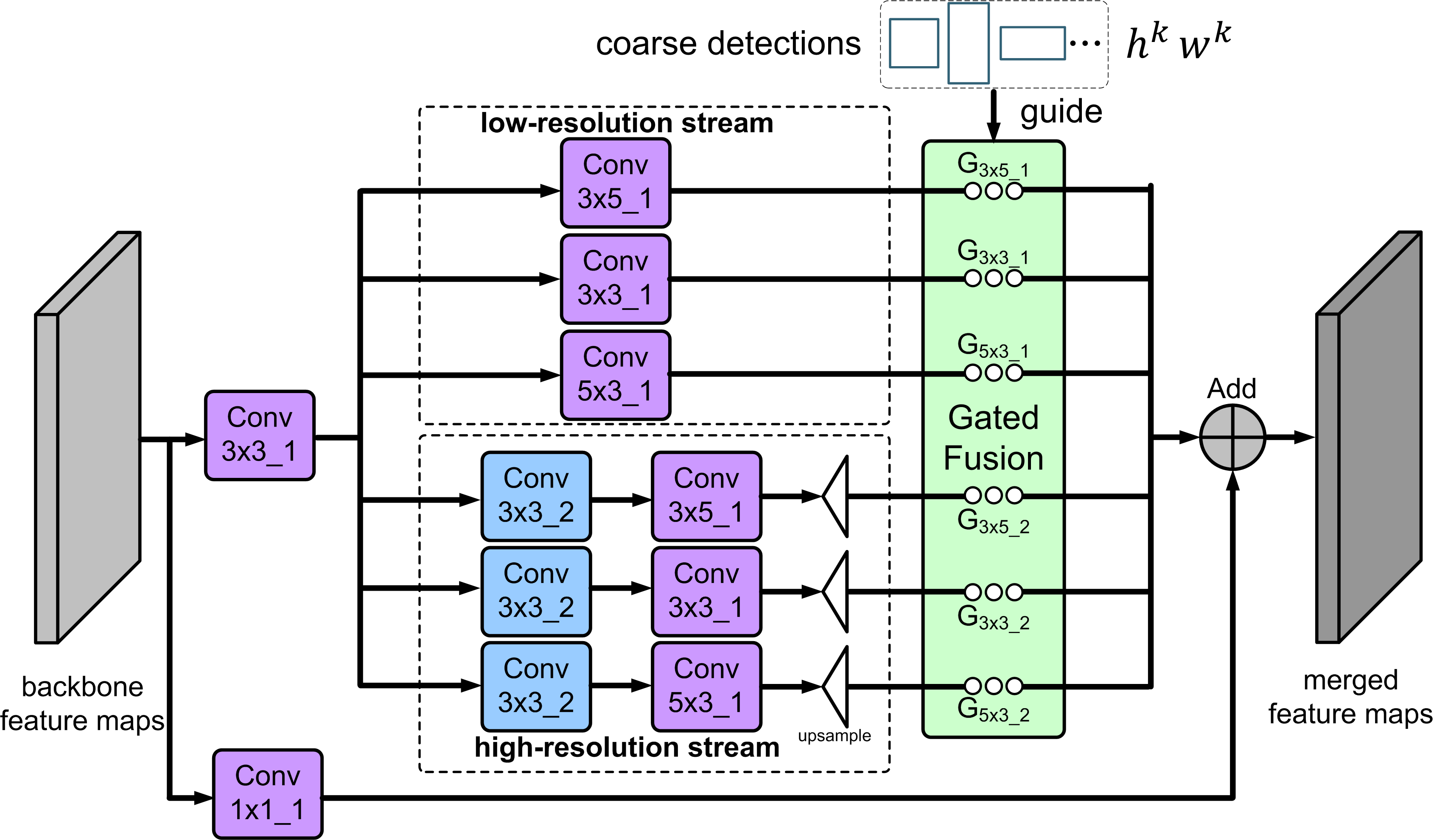}
\caption{Omni-scale gated fusion module for the FCOS detector \cite{DBLP:conf/iccv/TianSCH19}. ``3x3\_2'' in the blue rectangles denotes the $3\times3$ convolutional layer with stride $2$.}
\label{fig:gate}
\vspace{-4mm}
\end{figure}

{\noindent \textbf{Scale guidance.}} 
Followed by the multi-level feature maps $F^{k}$, we can predict the candidate object boxes $\tilde{b}^{k}$ by using a series of convolutional layers. According to \cite{DBLP:conf/cvpr/RezatofighiTGS019}, we use the cross-entropy Intersection over Union (IoU) loss \cite{DBLP:conf/mm/YuJWCH16} to regress the bounding boxes of objects in foreground pixels, \ie,
\begin{equation} \label{eq_candidate}
\mathcal{L}_\text{gui} = -\sum_k\sum_{(i,j)}\ln(\text{IoU}(\tilde{b}^k_{i,j}, b^k_{i,j})),
\end{equation}
where $\text{IoU}(\cdot,\cdot)$ is the function to calculate the IoU score between predicted box $\tilde{b}^{k}$ and ground-truth box $b^{k}$. For each pixel $(i, j)$ in the feature map, the corresponding box can be defined as a $4$-dimensional vector $b^{k}_{i,j}=(x_{t_{i,j}}, x_{b_{i,j}}, x_{l_{i,j}}, x_{r_{i,j}})$, denoting the distances between current location and the top, bottom, left and right bounds of ground-truth box respectively. Thus we can calculate the normalized object scale (\ie, width $w^k$ and height $h^k$) at each level as
\begin{equation}\label{eq_scale}
\left\{
\begin{aligned}
&w^k_{i,j} = (\tilde{x}_{r_{i,j}}+\tilde{x}_{l_{i,j}})/\text{stride}^k,\\
&h^k_{i,j} = (\tilde{x}_{b_{i,j}}+\tilde{x}_{t_{i,j}})/\text{stride}^k,\\
\end{aligned}
\right.
\end{equation}
where $\text{stride}^k$ denotes how many steps we are moving in each step in convolution\footnote{We have $\{(k,\textit{stride})|(3, 8), (4, 16), (5, 32), (6, 64), (7, 128)\}$.}. As defined in the FCOS detector \cite{DBLP:conf/iccv/TianSCH19}, the feature maps at each level are used to individually detect objects with different scales in the range $\{[-1, 64], [64, 128], [128, 256], [256, 512], [512, +\infty]\}$. Therefore, the majority of object scales is less than $8$, \ie, $w^k\leq8, h^k\leq8$. For simplicity, we omit the superscript $k$ and write $F$ for $F^k$ and $\tilde{b}$ for $\tilde{b}^k$ in the following sections.

{\noindent \textbf{Omni-scale gated fusion.}} 
To adapt to various scales of objects $\tilde{b}$ with different aspect ratios, we design the omni-scale gated fusion module that is composed of both \textit{low-resolution} and \textit{high-resolution} streams. 
As shown in Figure \ref{fig:gate}, the \textit{low-resolution} stream contains three parallel convolutional layers with different kernels $\omega\in\{3\times3,3\times5,5\times3\}$, which is used to extract features of small objects ($w^k\leq5, h^k\leq5$). 
In the \textit{high-resolution} stream, we first apply the $3\times3$ convolutional layers with stride $2$ to expand the receptive field and then convolutional layers with kernels $\omega$ to deal with large objects ($w^k>5, h^k>5$).

After that, we introduce the gate mask $G$ to weight each convolutional layer based on the predicted coarse boxes $\tilde{b}$, 
\begin{equation}\label{eq_mask}
G_{\omega} = \frac{\exp(\tau(o_{\omega}-\hat{o}))}{\sum_{\omega}\exp(\tau(o_{\omega}-\hat{o}))},
\end{equation}
where $\tau$ is the temperature factor. $o_{\omega}=\text{IoU}(\tilde{b}, \omega)$ denotes the overlap between the predicted box and the convolution kernel $\omega$. $\hat{o}$ is the maximal overlap among them. 
Finally, we can merge the pixel-level features to exploit the scale-wise representation of instances, \ie,
\begin{equation}\label{eq_feature}
M = \sum_{\omega}F_{3\times 3}\odot G_{\omega}+F_{1\times 1},
\end{equation}
where $\odot$ denotes the element-wise product. $F_\omega$ denotes the feature maps after the convolutional layer with kernel $\omega$.

{\noindent \textbf{Object detection.}} 
After obtaining the merged feature maps $M$, we predict the categories and bounding boxes of objects. In the FCOS network \cite{DBLP:conf/iccv/TianSCH19}, the object detection heads consist of classification, centerness and regression branches. The classification and centerness branches are optimized by the focal loss \cite{DBLP:conf/iccv/LinGGHD17} $\mathcal{L}_\text{cls}$ and cross-entropy loss \cite{DBLP:conf/iccv/TianSCH19} $\mathcal{L}_\text{ctr}$ respectively. The regression branch is optimized by the IoU loss \cite{DBLP:conf/mm/YuJWCH16} $\mathcal{L}_\text{reg}$. The loss function for object detection is defined as
\begin{equation} 
\mathcal{L}_\text{det} = \mathcal{L}_\text{cls} + \mathcal{L}_\text{ctr} + \mathcal{L}_\text{reg}.
\label{eqn_det_loss}
\end{equation}
More details of the above loss functions refer to \cite{DBLP:conf/iccv/TianSCH19}. 

\subsection{Multi-Granularity Discriminators}
As discussed in the introduction section, we apply the multi-granularity discriminators to identify whether the sample belongs to the source domain or target domain in different perspectives including pixels, instances and categories. The difference between two domains is reduced by the Gradient Reversal Layer (GRL) \cite{DBLP:conf/icml/GaninL15} that transfers reverse gradient when optimizing the object detection network. The discriminator consists of four stacked convolution-groupnorm-relu layers and an additional $3\times3$ convolutional layer. In the following, we describe our multi-granularity discriminators in detail.

{\noindent \textbf{Pixel-level and instance-level discriminators.}} 
Pixel- and instance-level discriminators are used to perform pixel-level and instance-level alignment of feature maps respectively. As shown in Figure \ref{fig:architecture}, given the input multi-level features $F$ and merged feature $M$, $L_\text{pix}$ and $L_\text{ins}$ employ the same loss function of the discriminator defined in Eq. \eqref{eq:dis_pix}. Similar to the previous work \cite{DBLP:conf/eccv/HsuTLY20}, we use the same loss function, denoted as $\mathcal{L}_\text{pix}$ and $\mathcal{L}_\text{ins}$. For example, the loss of the pixel-level discriminator $D^\text{pix}$ is defined as 
\begin{equation} \label{eq:dis_pix}
\begin{split}
\mathcal{L}_\text{pix} &= -\sum_{(i,j)} y^\text{pix}_{i,j}\log D^\text{pix}(F^s(i,j)) \\
       &+ (1-y^\text{pix}_{i,j})\log(1-D^\text{pix}(F^t(i,j))),
\end{split}
\end{equation}
where $F(i,j)$ denotes the feature at pixel $(i,j)$ in the feature map. We have the domain label $y^\text{pix}_{i,j}=1$ if the pixel is from source domain and $0$ otherwise. 

{\noindent \textbf{Category-level discriminator.}} 
As shown in Figure \ref{fig:discriminator}(c), our category-level discriminator is used to keep semantic consistency between different domain distribution. Specifically, we predict the category and domain labels of pixel $(i,j)$ in each image based on the output feature map $\hat{M}\in \mathbb{R}^{H\times W\times 2C}$, where $H$ and $W$ are the height and width respectively, $2C$ represents the total number of categories for source and target domains. 

Since there is no ground-truth to supervise the category-level discriminator, we assign pseudo labels to important samples with high confidence from object detection (see Sec. \ref{sec_detection}). In practice, given a batch of input images, we can output the category probability map $P$ using the object detection heads, and compute the maximum category probability over all levels $\bar{P}$. Let $\mathcal{S}$ denote the set of selected instances such that its probability is greater than the threshold, \ie, $\mathcal{S}=\{(i,j)|P_{i,j}>\theta_\text{cat}\bar{P}\}$. 
Then the instances in different categories are classified by Eq. \eqref{eq_dis_cls}, while the same category in two domains is aligned by Eq. \eqref{eq_sim_cls}:
\begin{itemize}
\item To keep \textbf{instance discriminability in different categories}, we separate the category distribution by using the following loss function:
\begin{equation} \label{eq_dis_cls}
\mathcal{L}_\text{dis} = -\frac{1}{|\mathcal{S}|}\sum_{(i,j)\in \mathcal{S}}\sum_{c=0}^{C-1}\hat{y}^\text{dis}_{i,j,c}\log(p^\text{dis}_{i,j,c}).
\end{equation}
By normalizing the confidence over the domain channel, $p_{i,j,c}^\text{dis}$ represents the probability of the $c$-th category of the pixel, \ie,
\begin{equation} \label{eq_cls_prob}
\small
p^\text{dis}_{i,j,c} = \frac{\exp{(\hat{M}_{i,j,2c}+\hat{M}_{i,j,2c+1})}}{\sum_{c=0}^{C-1}\exp{(\hat{M}_{i,j,2c}+\hat{M}_{i,j,2c+1})}},
\end{equation}
where $\hat{M}_{i,j,2c}$ and $\hat{M}_{i,j,2c+1}$ denote the confidence of the $c$-th category in source and target domains respectively (\textit{c.f.} Figure \ref{fig:discriminator}(c)). $\hat{y}^\text{dis}\in\mathbb{R}^{H\times W\times C}$ is the pseudo category label. We have $\hat{y}^\text{dis}_{i,j,c}=1$ if the instance at $(i,j)$ in $\hat{M}$ is an important one of the $c$-th category and $\hat{y}^\text{dis}_{i,j,c}=0$ otherwise.

\item \textbf{Category consistency in two domains.} After classifying instances of different categories, we need to further determine which domain the instance comes from. With the GRL \cite{DBLP:conf/icml/GaninL15}, the loss function can be written as
\begin{equation} \label{eq_sim_cls}
\mathcal{L}_\text{sim} = -\frac{1}{|\mathcal{S}|}\sum_{(i,j)\in\mathcal{S}}\sum_{m=0}^{2C-1}\hat{y}^\text{sim}_{i,j,m}\log(p^\text{sim}_{i,j,m}),
\end{equation}
where $y^\text{sim}\in\mathbb{R}^{H\times W\times 2C}$ is the pseudo domain label. Similarly, we have $\hat{y}^\text{sim}_{i,j,m}=1$ if the instance at $(i,j)$ in $\hat{M}$ is an important one of the $\lfloor\frac{m}{2}\rfloor$-th category in specific domain and $\hat{y}^\text{sim}_{i,j,m}=0$ otherwise. The domain probability $p^\text{sim}$ is
\begin{equation} \label{eq_sim_prob}
p^\text{sim}_{i,j,m} = \frac{\exp(\hat{M}_{i,j,m})}{\sum_{v=m-m\%2}^{m-m\%2+1}\exp(\hat{M}_{i,j,v})},
\end{equation}
where $\%$ is the remainder function. The loss function of the category-level discriminator $D^\text{cat}$ is written as
\begin{equation} \label{eq_cls}
\mathcal{L}_\text{cat} = \lambda_\text{dis}\mathcal{L}_\text{dis} + \lambda_\text{sim}\mathcal{L}_\text{sim},
\end{equation}
where $\lambda_\text{dis}$ and $\lambda_\text{sim}$ are the balancing factors.
\end{itemize}

\subsection{Overall Loss Function}
As discussed above, the omni-scale gated object detection network is supervised by $\mathcal{L}_\text{gui}$ and $\mathcal{L}_\text{det}$. Meanwhile, the multi-granularity discriminators are optimized in different granularities, including pixel-level $\mathcal{L}_\text{pix}$, instance-level $\mathcal{L}_\text{ins}$ and category-level $\mathcal{L}_\text{cat}$. In summary, the overall loss function is defined as
\begin{equation} \label{eq_overall}
\mathcal{L} = (\underbrace{\mathcal{L}_\text{gui} + \mathcal{L}_\text{det}}_{\text{object detection}}) + \alpha\cdot\underbrace{(\mathcal{L}_\text{pix} + \mathcal{L}_\text{ins} + \mathcal{L}_\text{cat})}_{\text{multi-granularity discriminators}},
\end{equation}
where $\alpha$ is the balancing factor between object detection and multi-granularity discriminators.

\begin{table*}[t]
\footnotesize
\setlength{\tabcolsep}{6pt}
\center
\begin{tabular}{cccccccccccc}
\hline
Method &Detector & Backbone & person & rider & car & truck & bus & train & mbike & bicycle & mAP  \\
\hline
Baseline &Faster-RCNN &VGG-16  & 17.8 & 23.6 & 27.1 & 11.9 & 23.8 & 9.1 & 14.4 & 22.8 & 18.8 \\
DAF \cite{DBLP:conf/cvpr/Chen0SDG18} &Faster-RCNN &VGG-16 &25.0 &31.0 &40.5 &22.1 &35.3 &20.2 &20.0 &27.1 &27.6 \\
SC-DA \cite{DBLP:conf/cvpr/ZhuPYSL19} &Faster-RCNN &VGG-16 & 33.5 & 38.0 & 48.5 & 26.5 & 39.0 & 23.3 & 28.0 & 33.6 & 33.8 \\
MAF \cite{DBLP:conf/iccv/HeZ19} &Faster-RCNN &VGG-16 &28.2 &39.5 &43.9 &23.8 &39.9 &33.3 &29.2 &33.9 &34.0 \\
SW-DA \cite{DBLP:conf/cvpr/SaitoUHS19} &Faster-RCNN &VGG-16 & 29.9 & 42.3 & 43.5 & 24.5 & 36.2 & 32.6 & 30.0 & 35.3 & 34.3 \\
DAM \cite{DBLP:conf/cvpr/KimJKCK19} &Faster-RCNN &VGG-16  & 30.8 & 40.5 & 44.3 & 27.2 & 38.4 & 34.5 & 28.4 & 32.2 & 34.6 \\
MOTR \cite{DBLP:conf/cvpr/CaiPNTDY19} &Faster-RCNN &ResNet-50 &30.6 &41.4 &44.0 &21.9 &38.6 &40.6 &28.3 &35.6 &35.1\\
CST \cite{DBLP:conf/eccv/ZhaoLXL20} &Faster-RCNN &VGG-16 &32.7 &44.4 &50.1 &21.7 &45.6 &25.4 &30.1 &36.8 &35.9\\
PD \cite{wu2021instance} &Faster-RCNN &VGG-16 &33.1 &43.4 &49.6 &22.0 &45.8 &32.0 &29.6 &37.1 &36.6\\
CDN \cite{DBLP:conf/eccv/SuWZTCQW20} &Faster-RCNN &VGG-16 &35.8 &45.7 &50.9 &30.1 &42.5 &29.8 &30.8 &36.5 &36.6\\
SFOD-Masoic-Defoggy \cite{DBLP:journals/corr/abs-2012-05400} &Faster-RCNN &VGG-16 &34.1 &44.4 &51.9 &30.4 &41.8 &25.7 &30.3 &37.2 &37.0\\
ATF \cite{DBLP:conf/eccv/HeZ20} &Faster-RCNN &VGG-16  & 34.6 & 46.5 & 49.2 & 23.5 & 43.1 & 29.2 & 33.2 & 39.0 & 37.3  \\
SW-Faster-ICR-CCR \cite{DBLP:conf/cvpr/XuZJW20} &Faster-RCNN &VGG-16 &32.9 &43.8 &49.2 &27.2 &45.1 &36.4 &30.3 &34.6 &37.4 \\
SCL \cite{DBLP:journals/corr/abs-1911-02559} &Faster-RCNN &VGG-16 &31.6 &44.0 &44.8 &30.4 &41.8 &40.7 &33.6 &36.2 &37.9 \\
CFFA \cite{DBLP:conf/cvpr/Zheng0LW20} &Faster-RCNN &VGG-16  & 43.2 & 37.4 & 52.1 & 34.7 & 34.0 &$\bf{46.9}$ & 29.9 & 30.8 & 38.6 \\
GPA \cite{DBLP:conf/cvpr/XuWNTZ20} &Faster-RCNN &ResNet-50 &32.9 &46.7 &54.1 &24.7 &45.7 &41.1 &32.4 &38.7 &39.5\\
SAPNet \cite{DBLP:conf/eccv/LiDZWLWZ20} &Faster-RCNN &VGG-16 &40.8 &46.7 &59.8 &24.3 &46.8 &37.5 &30.4 &40.7 &40.9 \\
UMT \cite{DBLP:conf/cvpr/Deng0CD21} &Faster-RCNN &VGG-16 &$\bf{56.5}$ &37.3 &48.6 &30.4 &33.0 &46.7 &$\bf{46.8}$ &34.1 &41.7 \\
MeGA-CDA \cite{DBLP:conf/cvpr/VSGOSP21} &Faster-RCNN &VGG-16 &37.7 &49.0 &52.4 &25.4 &49.2 &$\bf{46.9}$ &34.5 &39.0 &41.8 \\
CDG \cite{li2021category} &Faster-RCNN &VGG-16 &38.0 &47.4 &53.1 &34.2 &47.5 &41.1 &38.3 &38.9 &42.3 \\
ours &Faster-RCNN &VGG-16 &43.9 & $\bf{49.6}$ & $\bf{60.6}$ & 29.6 &$\bf{50.7}$ & 39.0 & 38.3 &$\bf{42.8}$  &$\bf{44.3}$\\
\hline
oracle &Faster-RCNN &VGG-16  &46.5 &51.3 &65.2 &32.6 &49.9 &34.2 &39.6 &45.8 &45.6  \\

\hline\hline
SST-AL \cite{DBLP:journals/corr/abs-2110-00249} &FCOS &- &45.1 & 47.4 &59.4 & 24.5 &50.0 &25.7 &26.0 &$\bf{38.7}$ & 39.6  \\
CFA \cite{DBLP:conf/eccv/HsuTLY20} &FCOS &VGG-16  &41.9 & 38.7 & 56.7 & 22.6 & 41.5 & 26.8 & 24.6 & 35.5 & 36.0  \\
CFA \cite{DBLP:conf/eccv/HsuTLY20} &FCOS &ResNet-101 &41.5 & 43.6 &57.1 & 29.4 &44.9 &39.7 &$\bf{29.0}$ & 36.1 & 40.2 \\
ours & FCOS  & VGG-16  & $\bf{45.7}$ &$\bf{47.5}$ &60.6 &$\bf{31.0}$ & 52.9 & 44.5 &$\bf{29.0}$ & 38.0  & 43.6\\
ours & FCOS  & ResNet-101  & 43.1 & 47.3 &$\bf{61.5}$ & 30.2 &$\bf{53.2}$ &$\bf{50.3}$ & 27.9 & 36.9 &$\bf{43.8}$\\
\hline
oracle &FCOS &VGG-16  & 50.1 &46.4 &68.0 &33.7 &54.5 & 38.7 & 30.7 & 39.7 & 45.2  \\
oracle &FCOS &ResNet-101 & 46.6 & 45.4 & 66.1 &33.6 & 54.1 & 62.9 & 29.0 & 37.1 & 46.9 \\
\hline
\end{tabular}
\caption{Weather adaptation detection results from Cityscapes to FoggyCityscapes.}
\label{tab_city_foggy}
\vspace{-4mm}
\end{table*}

\subsection{Implementation Details}\label{sec_impl}
{\noindent \textbf{Extension of our framework.}}
To extend our framework to Faster-RCNN \cite{DBLP:journals/pami/RenHG017}, we use the backbone features with the stride $16$ to collect the base feature maps $F$. Since Faster-RCNN \cite{DBLP:journals/pami/RenHG017} is a two-stage object detection method, we directly use the Region Proposal Network (RPN) to predict the coarse candidate boxes, supervised by the original RPN loss $\mathcal{L}_\text{gui} = \mathcal{L}_\text{rpn}$. Similarly, we use the classification and regression branches in the object detection heads to estimate the categories and bounding boxes of objects, defined as $\mathcal{L}_\text{det} = \mathcal{L}_\text{cls} + \mathcal{L}_\text{reg}$. Note that the RPN in Faster-RCNN \cite{DBLP:journals/pami/RenHG017} only predicts the top $K$ proposals. To fuse the feature maps with different convolutional layers, we first concatenate the feature map after each convolutional layer and then extract the features for each proposal by the ROIAlign operation. Finally, the merged features are determined by the corresponding object scales according to the RPN outputs. The detailed architecture of our method upon Faster-RCNN \cite{DBLP:journals/pami/RenHG017} can be found in the supplementary materials. 
 
{\noindent \textbf{Optimization strategy.}} 
We train the proposed network in two stages empirically. First, we disable the category-level discriminator and train the remaining network without multi-scale augmentation. Second, we fine-tune the whole network by adding the category-level discriminator and multi-scale augmentation. The model is trained with learning rate of $0.005$, momentum of $0.9$, and weight decay of $0.0001$. The balancing factors in Eq. \eqref{eq_cls} are set as $\lambda_\text{dis}=1.0, \lambda_\text{sim}=0.1$, and $\alpha$ in Eq. \eqref{eq_overall} is set as $0.1$.

\section{Experiments}
In this section, we compare our method upon different detectors (FCOS \cite{DBLP:conf/iccv/TianSCH19} and Faster-RCNN \cite{DBLP:journals/pami/RenHG017}) and backbones (VGG-16 \cite{DBLP:journals/corr/SimonyanZ14a} and ResNet-101 \cite{DBLP:conf/cvpr/HeZRS16}) with state-of-the-art domain adaptation methods. Moreover, we conduct a detailed ablation study to analyze the influence of important components in our model. Following \cite{DBLP:conf/cvpr/Chen0SDG18}, all the methods are evaluated using the mean average precisions (mAP) at the IoU threshold of $0.5$.

\subsection{Datasets}
Following~\cite{DBLP:conf/cvpr/Chen0SDG18}, the experiments are carried out on $7$ datasets including Cityscapes \cite{DBLP:conf/cvpr/CordtsORREBFRS16}, FoggyCityscapes \cite{DBLP:journals/ijcv/SakaridisDG18}, Sim10k \cite{DBLP:/conf/icra/driving17},  KITTI \cite{DBLP:/conf/cvpr/are12}, PASCAL VOC \cite{DBLP:journals/ijcv/EveringhamGWWZ10}, Clipart \cite{DBLP:conf/cvpr/InoueFYA18} and Watercolor \cite{DBLP:conf/cvpr/InoueFYA18}.

For weather adaptation, Cityscapes \cite{DBLP:conf/cvpr/CordtsORREBFRS16} is a dataset of outdoor street scenes in normal weather, including $2,975$ images for training set and $500$ images for validation set with $50$ different cities. As a natural target domain, FoggyCityscapes \cite{DBLP:journals/ijcv/SakaridisDG18} is a fog weather outdoor street scene dataset synthesized on Cityscapes \cite{DBLP:conf/cvpr/CordtsORREBFRS16}. 
Sim10k \cite{DBLP:/conf/icra/driving17} contains $10k$ images of the synthetic driving scene from the game video Grand Theft Auto V (GTA5). Thus the adaptation from Sim10k \cite{DBLP:/conf/icra/driving17} to Cityscapes \cite{DBLP:conf/cvpr/CordtsORREBFRS16} can be used for evaluation in synthetic-to-real adaptation.
Similar to Cityscapes \cite{DBLP:conf/cvpr/CordtsORREBFRS16}, KITTI \cite{DBLP:/conf/cvpr/are12} is another popular scene dataset with $7,481$ images in the training set. We verify the cross-camera adaptation ability from KITTI \cite{DBLP:/conf/cvpr/are12} to Cityscapes \cite{DBLP:conf/cvpr/CordtsORREBFRS16}. Note that only the class \textit{car} is considered in synthetic-to-real and cross-camera adaptations.
Besides, we evaluate domain adaptation methods on dissimilar domains, \ie, from PASCAL VOC \cite{DBLP:journals/ijcv/EveringhamGWWZ10} with real images to Clipart \cite{DBLP:conf/cvpr/InoueFYA18} and Watercolor \cite{DBLP:conf/cvpr/InoueFYA18} with artistic images. Note that we use $15k$ images in PASCAL VOC 2007 and 2012 training and validation sets as the source domain.
\begin{table}[t]
\small
\setlength{\tabcolsep}{3pt}
\begin{center}
\begin{tabular}{cccc}
\hline
Method &Detector & Backbone & mAP (\textit{car}) \\
\hline
Baseline &Faster-RCNN &VGG-16  & 30.1/30.2 \\
DAF \cite{DBLP:conf/cvpr/Chen0SDG18} &Faster-RCNN &VGG-16  & 39.0/38.5 \\
MAF \cite{DBLP:conf/iccv/HeZ19} &Faster-RCNN&VGG-16  & 41.1/41.0 \\
ATF \cite{DBLP:conf/eccv/HeZ20} &Faster-RCNN &VGG-16  & 42.8/42.1 \\
SC-DA \cite{DBLP:conf/cvpr/ZhuPYSL19} &Faster-RCNN &VGG-16  & 43.0/42.5 \\
UMT \cite{DBLP:conf/cvpr/Deng0CD21} &Faster-RCNN &VGG-16 & 43.1/- \\
SFOD-Mosaic \cite{DBLP:journals/corr/abs-2012-05400} &Faster-RCNN   &VGG-16   & 43.1/44.6 \\
CST \cite{DBLP:conf/eccv/ZhaoLXL20} &Faster-RCNN   &VGG-16   & 44.5/43.6 \\
MeGA-CDA \cite{DBLP:conf/cvpr/VSGOSP21} &Faster-RCNN &VGG-16 & 44.8/43.0 \\
SAPNet \cite{DBLP:conf/eccv/LiDZWLWZ20}&Faster-RCNN   &VGG-16   & 44.9/43.4 \\
CDN \cite{DBLP:conf/eccv/SuWZTCQW20} &Faster-RCNN   &VGG-16   & 49.3/44.9 \\
ours &Faster-RCNN &VGG-16  & $\bf{49.8}$/$\bf{45.2}$ \\
\hline
oracle &Faster-RCNN &VGG-16  & 66.9  \\
\hline\hline
SST-AL \cite{DBLP:journals/corr/abs-2110-00249} &FCOS &- & 51.8/45.6 \\
CFA \cite{DBLP:conf/eccv/HsuTLY20} &FCOS   &VGG-16   & 49.0/43.2 \\
CFA \cite{DBLP:conf/eccv/HsuTLY20} &FCOS  &ResNet-101  & 51.2/45.0 \\
ours&FCOS  &VGG-16  & $\bf{54.6}$/$\bf{48.5}$ \\
ours&FCOS  &ResNet-101   &  54.1/46.5 \\
\hline
oracle &FCOS  &VGG-16  &72.3   \\
oracle &FCOS  &ResNet-101  &71.3   \\
\hline
\end{tabular}
\end{center}
\vspace{-2mm}
\caption{Synthetic-to-Real/Cross-camera adaptation detection results from Sim10k/KITTI to Cityscapes.}
\label{tab_sim10k_city}
\vspace{-6mm}
\end{table}

\subsection{Result Analysis}
As presented in Table \ref{tab_city_foggy}, Table \ref{tab_sim10k_city}, and Table \ref{tab_pascal_clipart}, we compare our method with other state-of-the-art methods in various domain adaptation scenarios. Meanwhile, we provide the performance of the baseline Faster-RCNN \cite{DBLP:journals/pami/RenHG017} without adaptation. The ``oracle'' results indicate that we remove the discriminators in our network and then train and evaluate it on the target domain.

{\noindent \textbf{Cityscapes$\to$FoggyCityscapes.}} 
In Table \ref{tab_city_foggy}, we evaluate our method on weather adaptation datasets from Cityscapes \cite{DBLP:conf/cvpr/CordtsORREBFRS16} to FoggyCityscapes \cite{DBLP:journals/ijcv/SakaridisDG18}. By using FCOS \cite{DBLP:conf/iccv/TianSCH19}, our method achieves $3.6\%$ gain over the second best CFA \cite{DBLP:conf/eccv/HsuTLY20} with ResNet-101 backbone and more gain with VGG-16 backbone. By using Faster-RCNN \cite{DBLP:journals/pami/RenHG017}, our method still obtains better performance than the recent CDG \cite{li2021category}. In addition, our method performs slightly worse than the oracle results with different detection backbones, indicating the effectiveness of our model. 

{\noindent \textbf{Sim10k/KITTI$\to$Cityscapes.}} 
We provide the results on the synthetic-to-real adaptation datasets, where Sim10k \cite{DBLP:/conf/icra/driving17} is the source domain and Cityscapes \cite{DBLP:conf/cvpr/CordtsORREBFRS16} is the target domain. As shown in Table \ref{tab_sim10k_city}, our method achieves the best accuracy of $54.6\%$ with VGG-16 backbone and $54.1\%$ with ResNet-101 backbone respectively. Compared with CFA \cite{DBLP:conf/eccv/HsuTLY20} using FCOS \cite{DBLP:conf/iccv/TianSCH19}, our method obtains $5.6\%$ gain with VGG-16 and $2.9\%$ gain with ResNet-101 respectively. We notice there is a huge gap between the results of our method and oracle. This is because of a significant domain shift between synthetic Sim10k \cite{DBLP:/conf/icra/driving17} and real Cityscapes \cite{DBLP:conf/cvpr/CordtsORREBFRS16}.

We also present the comparison between our method and state-of-the-arts on cross-camera adaptation datasets. KITTI \cite{DBLP:/conf/cvpr/are12} is the source domain while Cityscapes \cite{DBLP:conf/cvpr/CordtsORREBFRS16} is the target domain. Compared with CFA \cite{DBLP:conf/eccv/HsuTLY20}, our method acquires $5.3\%$ and $1.5\%$ gains with VGG-16 \cite{DBLP:journals/corr/SimonyanZ14a} and ResNet-101 respectively, showing state-of-the-art performance using different backbones. 
It is worth mentioning that FCOS \cite{DBLP:conf/iccv/TianSCH19} with ResNet performs slightly inferior to the VGG counterpart. This is maybe because VGG features are more suitable than ResNet features for adaptation from Sim10k/KITTI to Cityscapes.

{\noindent \textbf{PASCAL VOC$\to$Clipart/Watercolor.}} 
In addition, we evaluate our method using Faster-RCNN \cite{DBLP:journals/pami/RenHG017} with ResNet-101 on real-to-artistic adaptation datasets from PASCAL VOC \cite{DBLP:journals/ijcv/EveringhamGWWZ10} to Clipart and Watercolor \cite{DBLP:conf/cvpr/InoueFYA18}. According to Table \ref{tab_pascal_clipart}, our method obtains the best mAP score of $44.8\%$ and $58.1\%$ on Clipart and Watercolor respectively, outperforming the second best UMT \cite{DBLP:conf/cvpr/Deng0CD21} slightly on Clipart. Since there exists a significant class imbalance (\ie, the label \textit{car}, \textit{cat} and \textit{dog} have much fewer images than other labels), our method even performs better than the oracle result on Watercolor \cite{DBLP:conf/cvpr/InoueFYA18}. By using our multi-granularity discriminators, the training samples in source domain can facilitate training an accurate detection network.

\begin{table}[t]
\small
\setlength{\tabcolsep}{6pt}
\begin{center}
\begin{tabular}{cccc}
\hline
Method & Detector &Backbone & mAP  \\
\hline
Baseline &Faster-RCNN &ResNet-101  & 27.8/44.6 \\
SW-DA \cite{DBLP:conf/cvpr/SaitoUHS19} &Faster-RCNN&ResNet-101  & 38.1/53.3 \\
SCL \cite{DBLP:journals/corr/abs-1911-02559} &Faster-RCNN&ResNet-101  & 41.5/55.2 \\
DBGL \cite{chen2021dual} &Faster-RCNN&ResNet-101  & 41.6/53.8 \\
ATF \cite{DBLP:conf/eccv/HeZ20} &Faster-RCNN&ResNet-101  & 42.1/54.9 \\
PD \cite{wu2021instance} &Faster-RCNN&ResNet-101  & 42.1/56.9 \\
SAPNet \cite{DBLP:conf/eccv/LiDZWLWZ20} &Faster-RCNN&ResNet-101  & 42.2/55.2 \\
UMT \cite{DBLP:conf/cvpr/Deng0CD21} &Faster-RCNN&ResNet-101  & 44.1/$\bf{58.1}$ \\
ours &Faster-RCNN &ResNet-101  & $\bf{44.8}$/$\bf{58.1}$  \\
\hline
oracle &Faster-RCNN&ResNet-101  & -/55.4 \\
\hline
\end{tabular}
\end{center}
\vspace{-2mm}
\caption{Real-to-Artistic adaptation detection results from PASCAL VOC to Clipart/Watercolor\protect\footnotemark.}
\label{tab_pascal_clipart}
\vspace{-6mm}
\end{table}
\footnotetext{There are no oracle results for the Clipart \cite{DBLP:conf/cvpr/InoueFYA18} dataset because we use all the images in the Clipart \cite{DBLP:conf/cvpr/InoueFYA18} dataset as the target domain.}
 
\subsection{Ablation Study}
\begin{figure*}[t]
\centering
\includegraphics[width=0.95\linewidth]{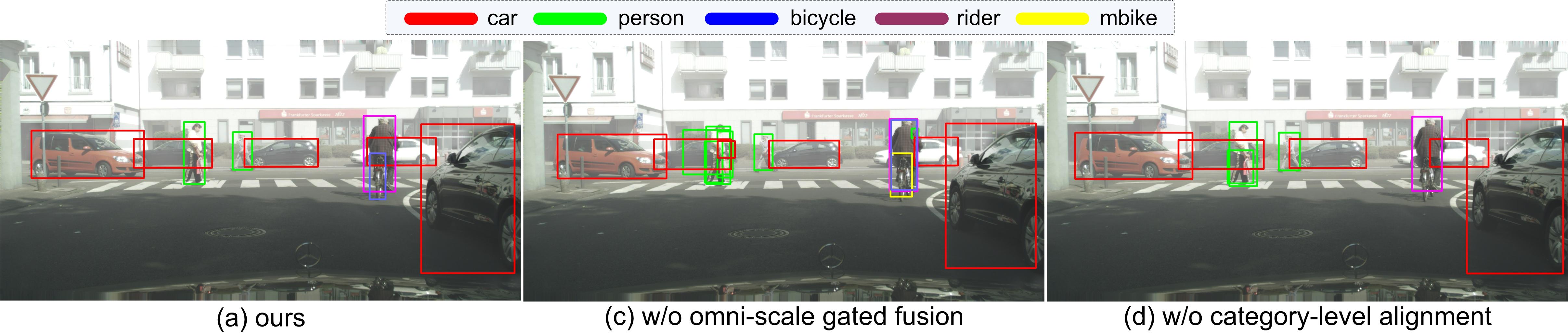}
\vspace{-2mm}
\caption{Visual comparison between our method and its variants.}
\label{fig:results}
\vspace{-4mm}
\end{figure*}

To study the effectiveness of important modules in our network, we conduct an ablation study on domain adaptation from Cityscapes \cite{DBLP:conf/cvpr/CordtsORREBFRS16} to FoggyCityscapes \cite{DBLP:journals/ijcv/SakaridisDG18}. We use FCOS \cite{DBLP:conf/iccv/TianSCH19} as the base detector with VGG-16 backbone for all the variants. As shown in Figure \ref{fig:results}, the visual results indicate that the proposed omni-scale gated fusion and category-level discriminator reduce false positives and negatives for object detection in adaptive domains.

{\noindent \textbf{Effectiveness of omni-scale gated fusion.}} 
To verify the ability of our method to deal with scale variations, we compare the mAP scores of our method and CFA \cite{DBLP:conf/eccv/HsuTLY20} in terms of object scale. According to COCO metrics, AP$^\text{S}$, AP$^\text{M}$ and AP$^\text{L}$ denote the mAP scores such that object area is in the range $[0,32^2], (32^2,96^2]$, and $(96^2, +\infty)$ respectively. 

In Table \ref{tab:gate}, the performance at all scales is considerably improved by using the omni-scale gated fusion in the object detection network compared with the baseline method, \ie, $39.3\%$ vs. $36.8\%$. If we remove the omni-scale gated fusion module before the object detection heads, the performance of ``ours (w/o gated fusion)'' is reduced by $2.3\%$. We also notice that our method performs better than CFA \cite{DBLP:conf/eccv/HsuTLY20} at all scales, especially at large scale. We speculate that the omni-scale gated fusion module can deal with various scales of objects to generate a more discriminative representation for object detection.

Moreover, we discuss the influence of coarse detection guidance for omni-scale gated fusion and object detection heads. If we use naive average fusion ``ours (w/ average fusion)'' or $1\times1$ convolution fusion ``ours (w/ conv fusion)'', the performance is lower than our method using the omni-scale gated fusion. It demonstrates that the coarse detection guidance is crucial to select the most plausible convolution for multi-scale feature aggregation.

\begin{table}[t]
  \centering
  \small      
  \setlength{\tabcolsep}{5.0pt}
    \begin{tabular}{c|c|ccc}
    \hline
    method & mAP & AP$^\text{S}$ & AP$^\text{M}$ & AP$^\text{L}$ \\
    \hline
    CFA \cite{DBLP:conf/eccv/HsuTLY20} & 36.0    & 8.3   & 36.7  & 61.6 \\
    ours (w/o all) & 36.8  & 7.2  & 37.7  & 64.1 \\
    ours (w/o category-level dis.) & 39.3  & 8.7   & 40.5  & 64.4 \\
    ours (w/o gated fusion) & 41.3  & 8.5  & 39.1  & 70.6 \\
    ours (w/ all) & $\bf{43.6}$  & $\bf{10.1}$  & $\bf{43.1}$  & $\bf{72.5}$ \\
    \hline
    ours (w/ average fusion) &42.1 &$\bf{11.5}$ &40.7 &68.9\\
    ours (w/ conv fusion) &41.5 &11.2 &40.1 &71.5 \\
    ours (w/ gated fusion) & $\bf{43.6}$  & 10.1  & $\bf{43.1}$  & $\bf{72.5}$ \\
    \hline
    \end{tabular}%
  \caption{Effectiveness of multi-scale object detection. Different variants of our method are contructed by removing important modules in the network. Moreover, two baseline fusion strategies are compared with our omni-scale gated fusion.}\label{tab:gate}
  \vspace{-4mm}
\end{table}%

{\noindent \textbf{Effectiveness of category-level discriminators.}} 
If we reduce the multi-granularity discriminators to classical discriminators by removing the category-level discriminator $D^\text{cat}$ in Eq. \eqref{eq_cls} from our method, we can observe the sharp drop of $4.3\%$ mAP of the baseline ($39.3$ vs. $43.6$), as presented in Table \ref{tab:gate}. It indicates the importance of our proposed discriminator. 

To further demonstrate the superiority of our category-level discriminator $D^\text{cat}$, we add three most related discriminators including $D^\text{ins}$ \cite{DBLP:conf/eccv/HsuTLY20}, $D^\text{grp}$ \cite{DBLP:conf/cvpr/HuKSC20}, and $D^\text{cls}$ \cite{DBLP:conf/eccv/WangSZD020} in our network with the baseline discriminators in Eq. \eqref{eq:dis_pix}. 
$D^\text{cen}$ \cite{DBLP:conf/eccv/HsuTLY20} considers the center-aware distribution alignment of pixel-level instances with its multi-scale extension. As shown in Figure \ref{fig:discriminator}(a), $D^\text{grp}$ \cite{DBLP:conf/cvpr/HuKSC20} utilizes category-level adversarial discriminator to decrease the differences within each category between source and target domain. As shown in Figure \ref{fig:discriminator}(b), $D^\text{cls}$ \cite{DBLP:conf/eccv/WangSZD020} expands the binary domain labels by using class information, and preserves intra-domain structures of source and target domains.
From Table \ref{tab_discriminators}, we obtain only $39.3\%$ mAP score by using the traditional pixel-level discriminator $D^\text{pix}$ in Eq. \eqref{eq:dis_pix}. By using either $D^\text{cen}$ or $D^\text{grp}$ in our method, the performance is improved with less than $2\%$ gain. Although $D^\text{cls}$ can further improve the performance slightly, our method achieves a considerable gain of near $5\%$. This is attributed to more instance discriminability in different categories over two domains in our method.

\begin{table}[t]
\small
\begin{center}
\setlength{\tabcolsep}{2pt}
\begin{tabular}{c|ccccc}
\hline
discriminator &baseline & $D^\text{cen}$ \cite{DBLP:conf/eccv/HsuTLY20} &$D^\text{grp}$ \cite{DBLP:conf/cvpr/HuKSC20}  &$D^\text{cls}$ \cite{DBLP:conf/eccv/WangSZD020} & $D^\text{cat}$ (ours) \\
\hline
mAP & 39.3 & 40.5 & 40.7 & 41.1 &$\bf{43.6}$  \\
\hline
\end{tabular}%
\end{center}
\vspace{-3mm}
\caption{Comparison between different discriminators. Existing discriminators including $D^\text{ins}$, $D^\text{grp}$, and $D^\text{cls}$ are added in our network with the baseline discriminators in Eq. \eqref{eq:dis_pix}.}
\label{tab_discriminators}
\vspace{-1mm}
\end{table}

{\bf Computational complexity.}
In addition, we provide the comparison of computational complexity between our method and other SOTA works in Table \ref{tab_speed}. Note that the discriminators are removed in testing phase for most methods except SCL \cite{DBLP:journals/corr/abs-1911-02559}. By using anchor-free FCOS, our multi-granularity alignment framework performs the best with reasonable increased complexity over its primary contender CFA \cite{DBLP:conf/eccv/HsuTLY20}; while our method has less parameters on top of anchor-based Faster-RCNN than two recent methods SCL \cite{DBLP:journals/corr/abs-1911-02559} and SAPNet \cite{DBLP:conf/eccv/LiDZWLWZ20}.

\begin{table}[t]
  \centering
  \scriptsize
  \setlength{\tabcolsep}{2pt}
   \begin{tabular}{c|cc|ccc}
    \hline
    method & CFA \cite{DBLP:conf/eccv/HsuTLY20} & ours & SCL \cite{DBLP:journals/corr/abs-1911-02559} &SAPNet \cite{DBLP:conf/eccv/LiDZWLWZ20} & ours \\
    \hline
    detector & FCOS & FCOS & Faster-RCNN & Faster-RCNN & Faster-RCNN \\
    \hline
    \# of Params (M) & 177   & 283   & 580   & 556   & 255 \\
    FPS   & 17.5  & 10.0  & 11.8  & 25.2  & 21.4 \\
    \hline
    \end{tabular}%
      \vspace{-1mm}
      \caption{Comparison of computational complexity.}
  \label{tab_speed}
  \vspace{-4mm}
\end{table}

\section{Conclusions}
In this work, we encode the multi-granularity dependencies among pixel-, instance- and category-level information to align the feature distribution of source domain and target domain in a more accurate way. Notably, the proposed omni-scale gated fusion module can exploit instance features among multi-scale feature maps with most plausible convolutions. Meanwhile, the multi-granularity discriminators can distinguish instances in different categories over two domains. The experiment shows the superiority of the above designs in our framework on top of different detectors and backbones for domain adaptive object detection.

{\noindent \textbf{Acknowledgement and Declaration of Conflicting Interests.}} 
Dr. Wu was supported by the Key Research Program of Frontier Sciences, CAS, Grant No. ZDBS-LY-JSC038. Dr. Zhang was supported by Youth Innovation Promotion Association, CAS (2020111). Dr. Du and his employer received no financial support for the research, authorship, and/or publication of this article.

{\small
\bibliographystyle{ieee_fullname}
\bibliography{egbib}
}

\end{document}